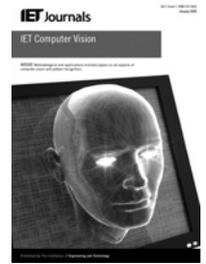

# Review of person re-identification techniques

Mohammad Ali Saghafi[1], Aini Hussain[1], Halimah Badioze Zaman[2], Mohamad Hanif Md. Saad[1]

[1]Faculty of Engineering and Built Environment, Universiti Kebangsaan Malaysia (UKM), Bangi, Malaysia
[2]Institute of Visual Informatics, Universiti Kebangsaan Malaysia (UKM), Bangi, Malaysia
E-mail: aini@eng.ukm.my

**Abstract:** Person re-identification across different surveillance cameras with disjoint fields of view has become one of the most interesting and challenging subjects in the area of intelligent video surveillance. Although several methods have been developed and proposed, certain limitations and unresolved issues remain. In all of the existing re-identification approaches, feature vectors are extracted from segmented still images or video frames. Different similarity or dissimilarity measures have been applied to these vectors. Some methods have used simple constant metrics, whereas others have utilised models to obtain optimised metrics. Some have created models based on local colour or texture information, and others have built models based on the gait of people. In general, the main objective of all these approaches is to achieve a higher-accuracy rate and lower-computational costs. This study summarises several developments in recent literature and discusses the various available methods used in person re-identification. Specifically, their advantages and disadvantages are mentioned and compared.

## 1 Introduction

One of the most important aspects of intelligent surveillance systems, which has been considered in the literature, is person re-identification, especially in cases in which more than one camera is used [1–3]. Re-identification is a pipelined process consisting of a series of image-processing techniques that finally indicate the same person who has appeared in different cameras. In identification, the entire process is performed under the same illumination, viewpoint and background conditions, but these conditions are uncontrolled in re-identification. Furthermore, in identification, a large number of samples are ready; meanwhile, in re-identification, one cannot expect to have seen the unknown person earlier. In fact, the combination of these uncontrolled conditions makes re-identification more difficult and at the same time more useful than identification in most cases. However, identification still has special applications in vision industries. The potential to make surveillance systems more operator-independent than before, and the need to address related issues that have not yet been solved, make re-identification an interesting subject of research.

### 1.1 What is people re-identification?

According to Paul McFedries [4], re-identification is the process of matching anonymous census data with the individuals who provided the data. Thus, the term 'people re-identification' can be defined as the process of matching individuals with a dataset, in which the samples contain different light, pose and background conditions from the query sample. The matching process could be considered as finding a person of interest among pre-recorded images, sequences of photos [5–7] or video frames [1, 2, 8–10] that track the individual in a network of cameras in real-time. Obviously, the latter is more challenging and has open unsolved issue and is being actively pursued by researchers worldwide.

### 1.2 Why is people re-identification significant?

Surveillance in public places is widely used to monitor various locations and the behaviour of people in those areas. Since events such as terrorist attacks in different public places have occurred more frequently in recent years, a growing need for video network systems to guarantee the safety of people has emerged. In addition, in public transport (airports, train stations or even inside trains and airplanes), intelligent surveillance has proven to be a useful tool for detecting and preventing potentially violent situations. Re-identification can also play a part in processes that is needed for activity analysis and event recognition or scene analysis. In an intelligent video surveillance system, a sequence of real-time video frames is grabbed from their source, normally closed circuit television (CCTV) and processed to extract the relevant information. Developing techniques that can process these frames to extract the desired data in an automatic and operator-independent manner is crucial for state-of-the-art applications of surveillance systems.

Today, the growth in the computational capabilities of intelligent systems, along with vision techniques, has provided new opportunities for the development of new approaches in video surveillance systems [1]. This includes automatic processing of video frames for surveillance purposes, such as segmentation, object detection, object recognition, tracking and classifying. One of the most important aspects in this area is person re-identification. As long as a person stays within a single camera's view, his position, as well as the lighting condition and background, is known to the system. However, problems arise in applications in which a network of cameras must be used when the person moves out of one camera's view and enters another. Although tracking a person within a single camera stream creates issues related to occlusion and is typically based on continuous user observations, multi-camera tracking raises the concern of uncovered areas where the user is not observed by any camera. Therefore how does the system know that the person seen in that camera was the same person seen earlier in another camera? This issue is known as a re-identification problem. It centres on the task of identifying people separated in time and location. The lack of spatial continuity for the information received from different camera observations makes person re-identification a complex problem.

The person re-identification problem has three aspects. First, there is a need to determine which parts should be segmented and compared (i.e. find the correspondences). Second, there is a need to generate invariant signatures for comparing the corresponding parts [11]. Third, an appropriate metric must be applied to compare the signatures. In most studies, the method is designed under the assumption that the appearance of the person remains unchanged [1, 12, 13] which seems sensible. Based on this assumption, local descriptors such as colour and texture can be considered to exploit the robust signatures of images. A collection of solutions have been used towards this objective, including gait [10, 14, 15], colour [11, 16–18], texture [19] and shape [20, 21] extraction methods. Each method has its own advantages and disadvantages. The methods are selected according to different scenarios, but each method has its own restrictions. For example, in appearance-based methods, person re-identification must deal with several challenges such as variations in the illumination conditions, poses (Fig. 1) and occlusions across time and cameras. In addition, different people may dress similarly. In gait-based methods [10], although there is no colour-constancy problem, the gait of a person appears to be different from different viewing angles and poses. Thus, the recognition rate diminishes when the individuals to be identified are viewed from different angles. The partial occlusions created by other people or objects also affect the gait-based methods.

There are two styles for writing surveys in the state-of-the-art re-identification literature, one is the method-based survey [21] and the other is phase-based [22–24]. In this survey, we have attempted to utilise a mixed style combining both styles. We have tried to discuss about the issues and their solutions more deeply compared with previous surveys. We have also separated different critical aspects in re-identification and discuss the most reliable methods that have been employed in these areas. As such, the review afforded in this current work is more extensive than the one reported earlier in [23].

The paper is organised as follows. In Section 2, the various issues regarding re-identification are explained. Section 3 describes the methods that have been used in re-identification. The most popular databases and evaluation metrics used for different methods are also reported in this section. In Section 4, we discuss the methods stated in the literature, and enumerate the pros and cons of these methods. The conclusion summarises the contents of this paper.

## 2 Issues regarding people re-identification

Problems related to re-identification make it more difficult than the identification task. Although some research has been done in this area, several problems have yet to be solved. These issues can generally be classified into two categories: (i) inter-camera and (ii) intra-camera. The problems may differ in different scenarios. For example, the considerations for re-identification in public places such as train stations or crowded shopping malls are different from those in a house. However, all applications have common problems.

To track the same person in different cameras, the system must be robust against illumination changes and outdoor clutter. Different camera viewpoints can also cause problems for methods that are based on the gait or shape of the moving person. Occlusion in public places is another issue that must be addressed. In methods based on the appearance of people, the clothing of the object shown from one camera to another should not be changed; otherwise, the system would fail. People will enter the camera's field of view with different poses; thus, for approaches that try to extract a model based on the movement of the person, changing the pose will create difficulties. To prevent failure, the designated methods must have the flexibility and capability to deal with these problems.

### 2.1 Inter-camera issues

Several issues (i.e. inter-camera issues) can cause problems in tracking people in a network of cameras with disjoint fields of view. Given that clothing is mostly characterised by its colour and texture, having a constant appearance of colour among all the cameras is important. The different illumination conditions that exist at different camera sites are problem to consider. Owing to bandwidth limits, images grabbed in a network of cameras will be compressed, causing unwanted noise to be added to these images [6]. Even if the cameras are from the same manufacturer, they have different features and so have differences in illumination. Another main problem is the different poses of humans of interest in different camera angles. This problem decreases the detection rate, especially in the gait-based methods. Several researchers have proposed the use of more robust methods to address this problem [12, 16, 25].

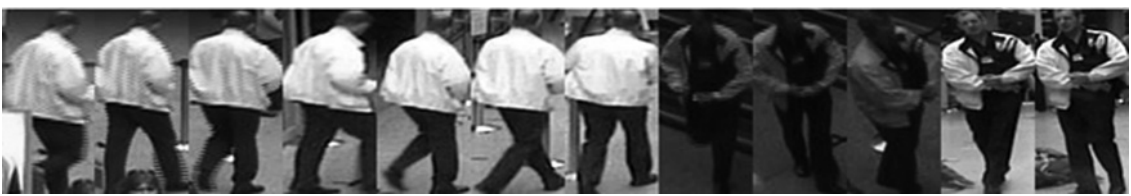

**Fig. 1** *Differences in poses and lighting conditions in four different cameras [26]*

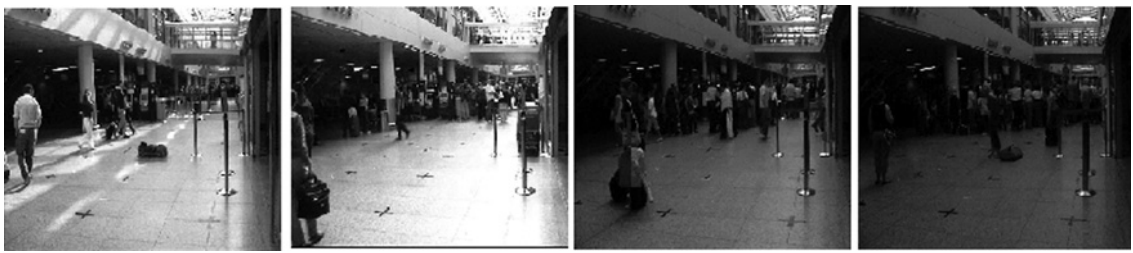

**Fig. 2** *Lighting conditions at different times of day [26]*

Tracking individuals between disjoint points of view is another problem in re-identification. Most methods rely on extracting the colour and texture features of an individual's clothing; however, offering methods that are invariant even with a rapid change of clothing would be better.

### 2.2 Intra-camera issues

Some of the problems to be addressed are related to varying light conditions at different times of the day (Fig. 2). In addition, most surveillance cameras are low-resolution cameras; hence, detection techniques that use methods that are dependent on the quality of the frames (e.g. face recognition methods [27]) can rarely be used. They have mostly been implemented and evaluated on local datasets rather than on standard famous datasets [25].

Occlusion (Fig. 3) in camera frames is another problem that creates difficulty in image segmentation (one of the steps in re-identification). As mentioned, the re-identification task is a pipelined process consisting of different processes, such as image segmentation, feature extraction and classification. Each of these processes represents a vast area of research in image processing and computer vision. Thus, there are specific considerations and issues related to them. We do not mention those concerns in this paper; we only consider the concatenation of these processes, which leads to the task of re-identification.

Table 1 presents the issues that must be overcome in re-identification. Some of these problems have been solved according to previous work, whereas others remain unsolved.

All of the methods that have been proposed for re-identification attempt to extract signatures (invariant features) from video frames and classify them in an appropriate manner to overcome the aforementioned problems. Thus far, there is no comprehensive framework reported in the literature that can cover all the issues related to re-identification, and each method can only partly cover the issues. In the following sections, different methods that have been investigated by several researchers are discussed.

**Table 1** Re-identification issues

| Types | Issues |
|---|---|
| inter-camera | illumination changes for different scenes; disjoint fields of view; different entrance poses in different cameras; similarity of clothing; rapid changes in person's clothing; and blurred images |
| intra-camera | background illumination changes; low resolution of CCTVs; and occlusion in frames |

## 3 Methods used for person re-identification

In this section, the most significant studies that have been done in the area of person re-identification in recent years are categorised and explained. Then, in Section 4, a summary of the methods, comparing their advantages and disadvantages, is presented. The backgrounds of most re-identification techniques in their present structures refer to multi-camera tracking approaches [28, 29], content-based image retrieval techniques [30, 31] and algorithms that have been used to extract the colour and texture information from still images to classify and label them among a large volume of raw data.

Generally, re-identification methods can be divided into two main groups. The first group includes methods that try to extract signatures from colour, texture and other appearance properties of frames. These are appearance-based methods. In contrast, others try to extract features from the gait and motion of persons of interest. These are gait-based methods, which are not popular yet because of the restrictions caused by different viewpoints or far-view camera frames in which the subject's gait is not clearly shown.

Whether the approach is appearance-based or gait-based, the re-identification consists of three main steps which are depicted in Fig. 4. The first step is to extract the blob of the person of interest from other parts of the image. The second step is to extract the signatures and the last step is to compare the extracted signatures and evaluate the similarities between the query and the gallery set.

### 3.1 Segmentation

Before we go through the feature extraction and classification stages, we briefly review the types of methods that are used in re-identification approaches as pre-processing, including human detection, background elimination and, in some work, shadow elimination. These pre-processing steps are not necessarily included in all re-identification approaches but applying them can increase the accuracy. The first step for re-identification after data acquisition is background elimination, which is needed to detect the person or region of interest.

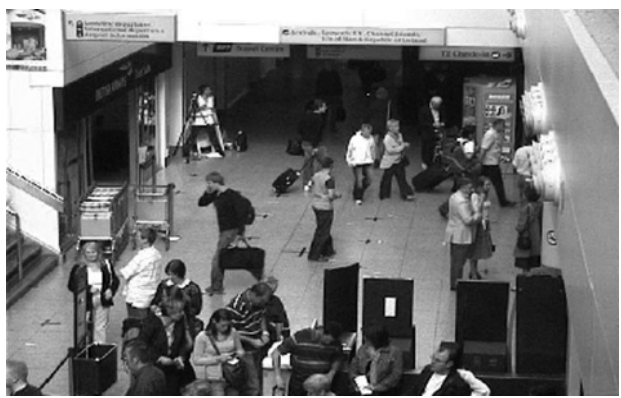

**Fig. 3** *Sample of occlusion in scene [26]*

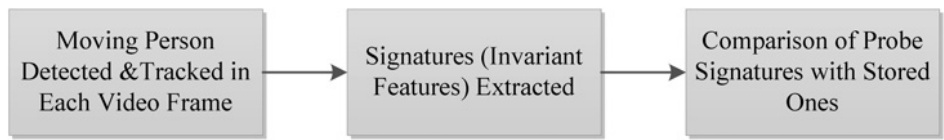

**Fig. 4** *Steps in re-identification*

*3.1.1 Background elimination:* Although there may be some approaches in which the background elimination is not suitable [32], in most of them it is necessary to remove the background to obtain better accuracy in the next stages. In re-identification, the same background may not exist for different frames because the data are grabbed from different cameras with different backgrounds. Thus, methods that use background subtraction with respect to a reference background frame are useless. Manual silhouette segmentation is the most naïve approach for background elimination method [33]. Gaussian mixture models (GMMs) [34] are widely used for background/foreground classifications [14, 35–38]. However, because GMM is sensitive to fast illumination variations, the background cannot be modelled accurately by using a limited number of components in cases with a high rate of illumination changes. Generative models like STEL [39] are also used in some approaches for background elimination in still images [16, 27, 40]. The main advantage of this method is that it can remove the background from the still images of a dataset with different backgrounds. However, it is time consuming and cannot be used in real-time applications. It also requires a fairly large number of samples for training. Bak *et al.* [41] used the probability density function of colour features of a target region to find the log-likelihood ratio of the foreground class. Gheissari *et al.* [11] used the maximum frequency image for background/foreground segmentation. Park *et al.* [6] also proposed a Gaussian background model using the two levels of pixels and image. In this approach, the mean and variance of the background model are updated recursively using temporal and spatial information. The methods used by Gheissari *et al.* and Park *et al.* are only useful in situations where we have sequences of frames, whereas in scenarios in which only still images of pedestrians are available, these methods cannot be utilised.

*3.1.2 Human detection:* Some approaches have preferred to use human detection and extract features from the bounding boxes that surround the human body. The histogram of oriented gradients (HOGs) proposed by Dalal and Triggs [42] is one of the most useful methods that have been applied for human detection [38, 41, 43–46] and even human body part detection [43, 47]. This method is suitable and can be utilised for cases in which sequences of video frames are not available, but needs training samples like structure element (STEL). It is reliable under different illumination conditions (mostly extracted from grey-scale images) and different scales (using a multi scale approach).

However, varying poses of humans may decrease the detection rate. A local binary pattern (LBP)-based detector was proposed by Corvee *et al.* [48] called simplified local binary pattern (SLBP), in which a set of 256 vector elements of an LBP texture descriptor was decreased to 16. In their work [48], they divided each cell into four parts, and then computed the mean intensities and mean differences of the parts to form the SLBP and used the AdaBoost classifier to train the features. To overcome the different scales, different sizes of cells were examined.

Goldmann [35] and Monari [49] used a pixel wise method proposed by Horpraser *et al.* [50] to detect persons in video streams. This algorithm closely mimics the human vision system in which the sensitivity to illumination is more than the sensitivity to the colour. In this method, the difference between pixel value of current image and background value in (red–blue–green) RGB colour space is decomposed into chromaticity and brightness components. The pixel is classified as foreground if only the chromaticity component exceeds a pre-defined threshold. For cases in which only the brightness component differs, the pixel is considered as shadow. In [51], Albiol *et al.* formed a height map based on the calculation of the pixels height from the ground. Next, to detect the moving persons or the blob, a threshold was applied on the height map image and followed analysis of connected component. In cases involving overlaps of two persons, watershed algorithm was used to split the blob. The criterion to split a blob is based on the existence of more than one local maximum (head position) on that blob.

One of the major concerns in human detection step for re-identification is the real-time implementation issue. Eisenbach and Kolarow [52] have used a real-time algorithm proposed by Wu *et al.* [53] which was capable of detecting human with 20 frames per second speed. This method uses census transform histograms visual descriptor which outperforms the HOG and LBP methods. This descriptor encodes the signs of comparisons of neighbouring pixels and composes a histogram of these codes. In contrast with HOG, the focus of this descriptor is on the contour information of images and only the sign information of neighbouring pixels is preserved while ignoring their magnitude information. The use of this human detection method in companion with an efficient, real-time colour-based human tracking method [54] empowered Eisnebach *et al.* work to track the persons through video frames in real-time.

In another work, Aziz *et al.* [55] have proposed one of the most applicable methods to detect human in crowd which was feasible in real-time application with only a short time delay. In this method, they performed background subtraction, and then used a particle filter to detect and track the heads and skeleton graphs in the video frames. This method was designed to work in a crowded scene which involved more than one person. In the case of occlusions where the body of two persons overlapped, the nearest head to the camera is kept and the other head is ignored.

The moving foreground of the silhouettes can also be tracked based on their spatial and colour probability distributions. Javed *et al.* as reported in [56] considered a Gaussian distribution of spatial probability density function for moving persons in consequent frames and used normalised colour histograms of foreground pixels as the objects' colour distribution. A foreground pixel which had the maximum colour and spatial product value voted for an object label. In the next level, a foreground region in the current frame is assigned to an object when most of its pixels (above a threshold) had voted for that object. In the case of partial occlusion, the position of partially occluded object was indicated based on the mean and variance of the pixels that voted for that object label.

*3.1.3 Shadow suppression:* Sometimes, the shadows are not eliminated in the background subtraction step. Thus, some methods are being used to remove the remaining shadows. Roy *et al.* [14] used the method in [57], in which after the background subtraction the angle between the background pixel value and the foreground pixel is compared with a threshold to decide whether or not the pixel belongs to a shadow. Park *et al.* [6] also used the same proposed Gaussian model that they had used for background subtraction by applying it on the foreground pixels in Hue Saturation Value (HSV) space. The subtraction was first performed on V and then on the H and S values. In [10], shadows were detected if the difference between the pixel value and the expected value of the Gaussian background model was within two thresholds.

## 3.2 Spatial features in re-identification

The models that are created to describe the appearance or gait can be extracted using a holistic description of an individual [58] or by a part-based or region-based description of that individual [16, 21, 59]. In both appearance-based and motion-based approaches, using the intermediate step of extracting the spatial information helps to extract more robust features and finally obtain a better re-identification rate. The partitioning can be done based on fixed proportions of the bounding box around the person of interest [59], but this cannot properly separate the regions and portions.

*3.2.1 Symmetrical and asymmetrical axes:* One of the most applicable partitioning algorithms, which has been utilised in many approaches [27, 52, 60] was proposed by Farenzena *et al.* [16]. In this algorithm, three main body regions are divided by two horizontal asymmetry axes corresponding to the head, torso and legs. This division is based on the maximum difference between the number of pixels in two moving rectangles, which sweeps the image to divide the head and torso and also the maximum colour difference in these bounding boxes to divide the torso and legs. In the last two regions, a vertical axis of the appearance symmetry is estimated. The use of symmetrical axes (by giving the pixels weights based on their distance from the axes) has a significant improvement to make the method pose independent. Fig. 5 shows body separation by this method. In [59], two rectangles like that in [16] were used to scan the image and find the best separating line between the torso and the legs. The decision here to find the maximum colour dissimilarity of the torso and legs was made based on the Bhattacharyya coefficients of the histograms of two rectangles.

*3.2.2 Spatiotemporal segmentation:* Gheissari *et al.* [11] proposed an algorithm that segments the silhouette based on its salient edges. It is robust against cloth wrinkles and temporary edges caused by changing illumination conditions in different frames. The significance of this method is that it groups pixels based on their fabric. In this method, an over segmentation is first performed using watershed algorithm, which results in a set of contiguous regions, as shown in Fig. 6. In the next step, a graph $G = \{V, E\}$ with spatial and temporal edges is defined. If two regions in one frame share a common boundary their corresponding edge is $e_{i,i'}^{t,t}$ and if two regions in consequent frames are indicated as corresponding ones their edge is $e_{i,i'}^{t,t+1}$. The correspondence between two regions in two consequent frames is detected by the frequency image of frames. Finally, a graph-based partitioning algorithm is used to group the connected regions (temporally and spatially), where their inter-class variations are less than their intra-class variations.

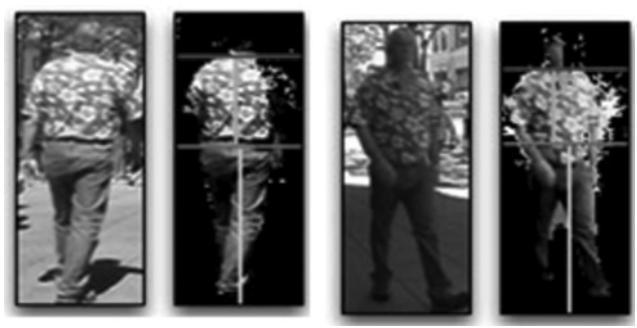

**Fig. 5** *Body segmentation using symmetry and asymmetry axes [16]*

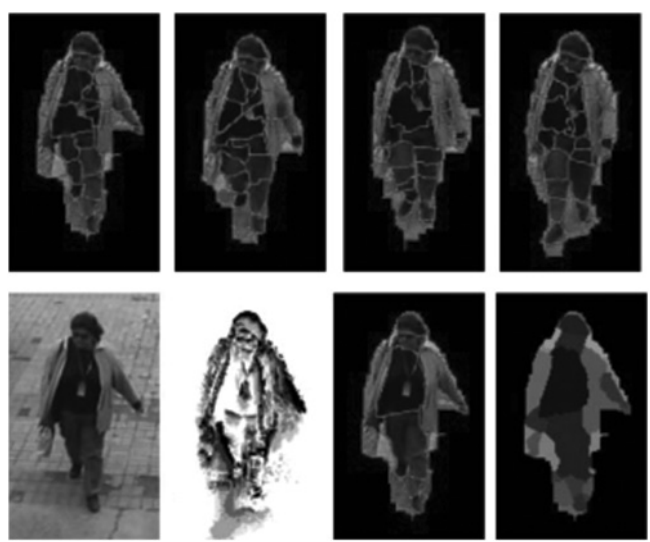

**Fig. 6** *Spatiotemporal segmentation [11]*

*3.2.3 HOG as body part detector:* As previously mentioned, in addition to using HOG for human detection it can be used to detect body parts. Bak *et al.* [43] and Bedagkar-Gala and Shah [47] used HOG as part detector as shown in Fig. 7. The idea is the same as using it for human detection and the system must be trained by negative and positive samples for each body part. This type of segmentation allows the algorithm to compare corresponding parts with each other in the classification stage which will decrease the computation cost.

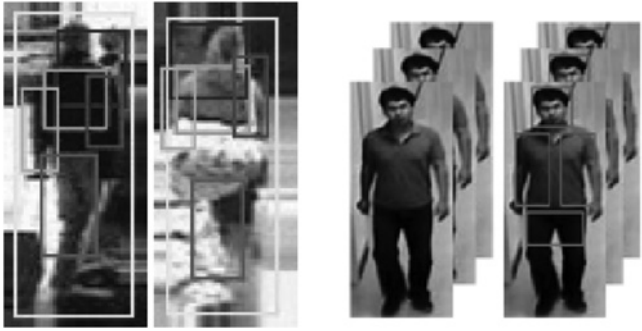

**Fig. 7** *HOG body part detector [43, 47]*

*3.2.4 Group representation of people for re-identification:* In contrast to most of the approaches used for re-identification which emphasise detecting and re-identifying an individual person of interest, Zheng et al. [20] proposed two spatial descriptors in which associated groups of people are considered as visual contexts. These descriptors are invariant from the rotations and swaps that occur in associating groups and are definitely invariant from scene occlusion. In fact, rather than re-identifying pre-viewed persons individually, these spatial descriptors try to re-identify pre-viewed groups of people. The SIFT-RGB features of the image are extracted and classified into $n$ visual words of $w_1, \ldots, w_n$. Then, the pixel values are replaced with their corresponding visual words, and the image is divided into $l$ non-overlapped regions ($p_1, \ldots, p_l$) that expand from the centre of the image which is depicted in Fig. 8. For each region $p_i$, a histogram $h_i$ is built, where $h_i(a)$ means the frequency of occurrence of visual word $w_a$ in that ring. An intra-ratio occurrence index $h_i(a, b)$ is also defined which indicates the ratio of the frequency of occurrence of $w_a$ to $w_a + w_b$. This is defined as follows

$$h_i(a, b) = \frac{h_i(a)}{h_i(a) + h_i(b) + \varepsilon} \quad (1)$$

To obtain inter-ratio occurrence indices, $g_i$ and $s_i$ are first defined as follows

$$g_i = \sum_{j=1}^{i-1} h_j, \quad s_i = \sum_{j=i+1}^{l} h_j \quad (2)$$

Finally, $G_i(a, b)$ and $S_i(a, b)$ are defined as inter-ratio occurrence indices

$$G_i(a, b) = \frac{g_i(a)}{g_i(a) + g_i(b) + \varepsilon},$$
$$S_i(a, b) = \frac{s_i(a)}{s_i(a) + s_i(b) + \varepsilon} \quad (3)$$

Therefore, for each region $p_i$, the centre rectangular ring ratio-occurrence (CRRRO) descriptor will be defined as $T_r^i = \{H_i, G_i, S_i\}$ and the whole image will be described by $\{T_r^i\}_{i=1}^{l}$. Fig. 9 shows how CRRRO is helpful for extracting the inter-person spatial information of groups of people.

Another spatial descriptor that is defined in this approach is the block based ratio-occurrence (BRO) descriptor. This descriptor is defined to extract the likely local patch information from individuals, as it can be seen in Fig. 8 (right). The image is divided into grid blocks, and BRO is defined inside each block. Each block $B_i$ is divided into sub-blocks $SB_{\gamma i}$ ($\gamma = 1$). In fact, this descriptor copes with the non-rotational position changes of a person which CRRRO cannot do. A complementary sub-region $SB_{\gamma i+1}$ is also considered to cover the other visually similar blocks in the same group of people. Similar to CRRRO, the index $H_j^i$ is defined in this descriptor between visual words in each region $SB_i$, but an extra index $O_j^i$ is defined here to explore the inter-ratio occurrence of $SB_i$ and other block regions

$$O_1^i(a, b) = \frac{t_i(a)}{t_i(a) + z_i(b) + \varepsilon},$$
$$O_2^i(a, b) = \frac{z_i(a)}{t_i(a) + z_i(b) + \varepsilon} \quad (4)$$

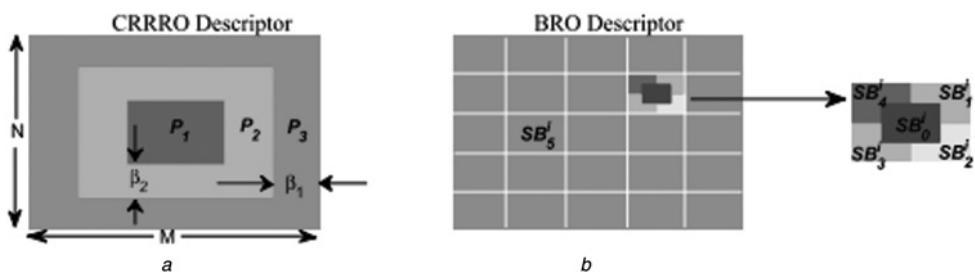

**Fig. 8** *CRRRO (left) and BRO (right) descriptors [20]*

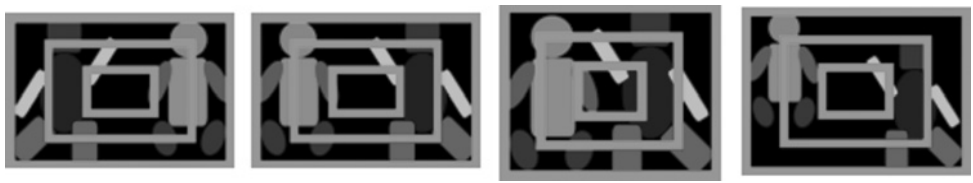

**Fig. 9** *Inter-spatial information extraction in CRRRO [20]*

where $t_i$ and $z_i$ are visual word histograms of $B_i$ and complementary image region $SB_{\gamma i+1}$. Thus, the BRO will be represented by $T_b^i = \{H_j^i\}_{j=0}^{4\gamma+1} \cup \{O_j^i\}_{j=1}^2$, $i = 1, \ldots, m$ where $m$ is the number of blocks.

### 3.3 Appearance-based models

According to the literature, appearance-based methods are more suitable in re-identification because of the short time required by the entire process. Although varying pose and illumination issues have a direct effect on the appearance features (like colour or texture) of different images, the availability and discrimination potential of appearance-based features are the likely cause of using them in most of the works on re-identification. In this section, models based on the appearance features of images will be further discussed.

*3.3.1 Colour histograms:* Colour histograms are the most popular tools used to describe the occurrence frequency of colours in an image. In re-identification, the holistic representation of a scene is not applicable and effective. Thus, histograms are preferably extracted from the segmented parts and regions [11, 12, 16, 27, 61]. The colour histograms are defined in different colour spaces. RGB colour histograms [12, 61], HSV colour histograms [16, 27, 32, 62] or LAB colour space histograms [62, 63] are examples of using different colour spaces to construct histograms. Among these colour spaces, HSV channels are the more robust against illumination changes. The luminance and chromatic channels are also separated from each other in the LAB colour space and can therefore be helpful to overcome the effects of the varying illumination of different frames. The main disadvantage of histograms is the lack of geometric and spatial information, which is necessary in re-identification applications. To add spatial information to histograms, the silhouette can be divided into horizontal stripes, where a colour-position histogram is defined for each stripe [57, 58]. Spatiograms can also provide complementary spatial information by adding higher-order spatial moments to histograms [2].

D'Angelo and Dugelay [63] proposed a probabilistic representation of histograms called probabilistic colour histogram (PCH), in which the colours were quantised into 11 culture colours using a fuzzy *k*-nearest neighbour algorithm. In this approach, a data element can belong to more than one cluster. A membership vector $\mathbf{u}(n) = \{u_1(n), u_2(n), \ldots, u_C(n)\}$ is defined for each pixel which indicates the degree of the association of pixel $n$ in all $C = 11$ clusters. Then, for each segmented part PCH is defined as a vector of $\mathbf{H}(X)$

$$\mathbf{H}(X) = \{H_1(X), H_2(X), \ldots, H_{11}(X)\},$$
$$H_c(X) = \frac{1}{N} \sum_{n=1}^{N} u_c(X_n) \quad (5)$$

where $N$ is the number of total pixels in that segment. The fuzzy clustering used in this method causes the pixel to belong to more than one cluster. The quantisation of colours into 11 culture colours and the fuzzy nature of histogram make this method more reliable against illumination changes than normal histograms. However, the histogram does not contain spatial information. In addition, the method has not been compared with any other method to provide information on any improvement in results compared with normal histograms. The fuzzy space colour histogram (FSCH) that was recently proposed by Xiang *et al.* [37] contained both colour and space information. In this approach, the usual three-dimensional (3D) histogram (in RGB space) was replaced by a 5D fuzzy histogram $\tilde{P}(R, G, B, x, y)$, which included the pixel geometry. A membership function $w_i(x)$ was also defined so that each pixel belonged to two neighbouring bins at each dimension. In the implementation stage for re-identification, the authors reduced the dimensionality of the histogram to 4D by removing the $x$ dimension to make the histogram robust against pose variations.

*3.3.2 Colour context people descriptor:* The idea of colour context people descriptor (CCPD) proposed by Khan *et al.* [59] was inspired by the shape context structure in Belongie *et al.* [64]. In this approach, the shape context structure is placed in the centre of the segmented object (which is torso or legs). Then, based on the pixels' radial and angular bins, a colour histogram is generated. The 3D structure of CCPD is shown in Fig. 10.

Depending on the pose, the legs may contribute more or less colour information to the histogram. Therefore the bottom histogram for the same person will be different from pose to pose. Thus, it is important to ignore the background pixels and only consider the legs' pixels to make the descriptor more discriminative. A back projection algorithm [65] is used to identify the pixels that represent the legs. Colour histograms of the both top and bottom rectangular areas are created, and the Bhattacharyya coefficient [66] is computed to match the histograms. Applying the CCPD on the whole torso region enters some unwanted pixels from the background in histogram computation. To improve this problem, it is better to apply a background/foreground segmentation algorithm on detected person bounding box and then apply CCPD or to apply it on smaller extracted patches from the torso region.

*3.3.3 MPEG7 colour descriptors for surveillance retrieval:* MPEG7 colour descriptors have generally been used in image retrieval applications [67, 68]; however, a number of researchers have used these descriptors specifically for re-identification purposes. The visual descriptors in the MPEG7 standard use both colour and spatial information, and most of them are invariant against scale, rotation and translation, which is why MPEG7 is considered to be potentially capable descriptor for re-identification. Annesley *et al.* [33] used MPEG7 colour descriptors (dominant colour, colour layout, scalable colour and colour structure) to re-identify a person in a dataset grabbed by different cameras at different times. They collected a set of image sequences of pedestrians entering and leaving a room, viewed by two cameras, as the test set (data are from medium and far-view image sequences). A dataset was generated from the top and bottom clothing components of each individual. They evaluated how this kind of segmentation affects the retrieval accuracy rate of the system. In addition, they investigated the effect of combining colour descriptors to improve the retrieval accuracy rate. However, in multiple camera datasets, the MPEG7 colour descriptors do not outperform the simple (R, G, B) mean description of foreground data because of the lack of colour constancy.

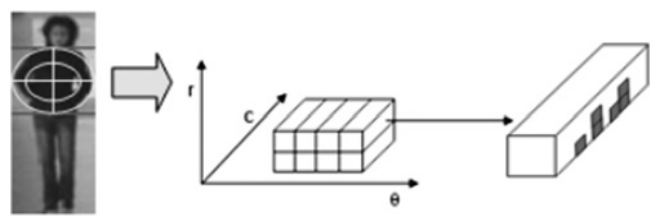

**Fig. 10** *CCPD [59]*

In another work, Bak *et al.* [41] used the dominant colour descriptor (DCD) to extract robust signatures from segmented images. In this approach, a human detection algorithm was used to find people in video sequences, and then an individual was tracked through several frames to generate a human signature. The DCD signature was created by extracting the dominant colours of upper and lower-body parts. These two sets were combined using the AdaBoost scheme to capture different appearances corresponding to one individual. The method used cross-correlation model functions to be robust against differences in illumination and pose, and to handle inter-camera colour calibration.

*3.3.4 Interest point detectors and descriptors in re-identification:* Gheissari *et al.* [11] proposed a method based on generating a large amount of interest points (colour and structure information) around regions with high-information contents (Fig. 11). The Hessian affine invariant operator [69] was used to nominate interest regions. The HSV histogram and 'edgel' histogram are two colour and structural features exploited from these regions to be compared. In [60], Martinel *et al.* used scale-invariant feature transform (SIFT) [70] interest points as the centres of circular regions, and a Gaussian function was used to construct a weighted colour histogram from the interest regions. The SIFT interest points are 3D histograms of the location and gradient orientation of the pixels. The gradient location and orientation histogram (GLOH) [71] is another descriptor, which is actually an extension of the SIFT descriptor to improve its distinctiveness [72]. In another attempt, Hamdoun *et al.* [3, 58] proposed a method based on harvesting SIFT-like interest point descriptors from different frames of a video sequence. In contrast to the method mentioned in [11] where matches are done image-to-image, this method exploits a sequence of images. It generates a more dynamic and multi-view descriptor than the use of a single image. In the learning phase of this algorithm, the given object, person, or car is tracked in one camera to extract interest points and descriptors to build the model. The interest point detection and descriptor computation is conducted using the 'key points' functions available in the Camellia image processing library, which was inspired by speeded up robust feature (SURF) [73], but is even faster because the detector mostly relies on integral images to approximate the determinant of the Hessian matrix. De Oliveira and De Souza Pio [74] also used the SURF descriptor to locate interest regions, including the HSV histogram of the points and saved it as a compact signature. The correlation of compact signatures was then computed to find the best matches. The main advantage of using interest points for detection and description is their invariance to illumination and partial invariance to pose changes. However, the redundancy of interest points is not desirable and must be limited. The other issue which must be taken into account is that the interest point detectors are sensitive to edges so their performance on the silhouette edges may be decreased.

*3.3.5 Covariance descriptor for re-identification:* The insensitivity to noise and invariance to the identical shifting of colours make the covariance descriptor suitable for re-identification [43, 52, 75–78]. If we consider $R$ as a segmented part of an image $I$, the covariance descriptor of region $R$ will be defined as a $d \times d$ dimensional covariance matrix as follows [18]

$$C_R = \frac{1}{n-1} \sum_{k=1}^{n} (f_k - \mu)(f_k - \mu)^T \quad (6)$$

where $\{f_k\}_{k=1,\ldots,n}$ are the $d$-dimensional feature points of region $R$ with $n$ number of points of region $R$ and the mean $\mu$ for the region points. The feature vector in the covariance descriptor can contain the colour, gradient or spatial derivatives of points. Bak *et al.* [43] proposed the spatial covariance regions (SCRs) descriptor, in which the physical locations of the points, along with their RGB colours, the gradient's magnitudes and their orientations were used to construct the feature vector. This model handles differences in the illumination, pose and camera parameters. In this person re-identification approach, the human detector and respective body parts detector based on the HOG are applied to establish the correspondence between body parts. Then, the covariance descriptor is offered to identify the similarity between corresponding body parts. Finally, an advantage of the concept of spatial pyramid matching [79] is used to design a new dissimilarity measure between human signatures. Hirzer *et al.* [78] used the covariance descriptor of the horizontal stripes of an image patch. The feature vector in their descriptor contained $y$ position, LAB colour channels and vertical/horizontal derivatives of the luminosity channel. The irrelevancy to the $x$ axes made the descriptor robust against pose changes, but the naïve horizontal segmentation of the images made it less discriminative than the part-based segmentation in [43] or the dense grid segmentation in [75]. The feature vector can consist of Gabor and LBP texture features [19]. A Gabor mask with orientation '0' was used to make the descriptor invariant against pose changes, and the LBP can provide invariance to grey level changes. In other studies of these authors [48, 75] they proposed the mean Riemannian covariance (MRC) descriptor, which was the temporal mean of the covariance matrices of overlapped regions. The feature vector of the covariance descriptor was almost the same as that in their previous work [43]. The covariance descriptors are robust through rotation and illumination changes, and dense representation (overlapped regions) makes this descriptor robust to partial occlusion. However, it must be noted that the covariance descriptors are not defined in Euclidean space and do not have additive structure. Thus, every operation, like the mean or variance, must be specially treated, which leads to greater computation cost. One solution is to compare only the means of the covariance matrices of corresponding parts instead of the whole descriptor [76].

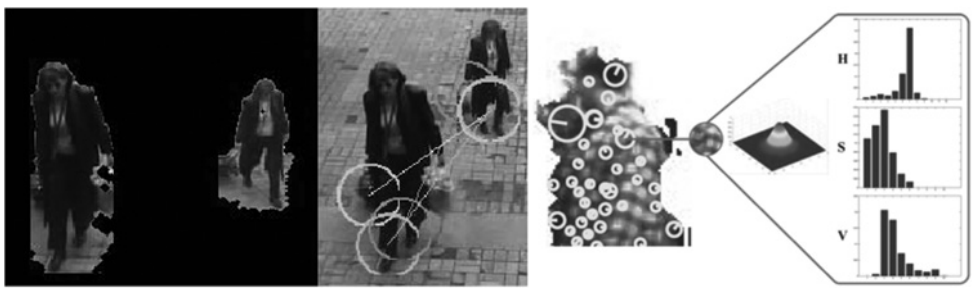

**Fig. 11** *Interest point detectors (SIFT detector) from [11] and [60]*

*3.3.6 Textural descriptors in re-identification:*
Texture features play complementary role in constructing appearance-based models. In re-identification, they must be combined with colour features to improve the performance. The only texture-based descriptors are not very effective in re-identification [19]. The recurrent high-structured patches (RHSP) descriptor proposed by Farenzena *et al.* [16] is one of the most applicable texture descriptors that are utilised in some other works [38, 52]. The descriptor is based on selecting patches from the foreground and ignoring the one with low-structural information by thresholding their entropy. Some transformation was done on pruned patches to evaluate their invariance through geometric variations, which were then mixed to construct the RHSP. This texture descriptor is invariant through pose and rotation changes, but needs images with at least a medium resolution to be applicable. Fig. 12 shows the extracted RHSP texture features from a torso.

Gabor [80] and Schmid [81] filters are mostly applied on luminance channels. These filters are rotation invariant. Thus, they create features that are pose and view point invariant for re-identification.

Co-occurrence matrices also have been used for texture description [35]. This descriptor is constructed from squared matrices, which provide information about the neighbouring pixels' relativity. The joint probabilities of neighbour pixels $P(i, j)$ inside the square matrix with dimension $(N \times N)$ can be described as

$$P(i,j) = \sum_{x=1}^{N} \sum_{y=1}^{N} \begin{cases} 1, & \text{if } I(x, y) = i, \ I(x + \Delta x, y + \Delta y) = j \\ 0, & \text{otherwise} \end{cases}$$
(7)

where $(\Delta x, \Delta y)$ is the distance between the pixel of interest and its neighbour and $I(x, y)$ is the pixel value at point $(x, y)$. Most of the texture descriptors are defined on grey level channels and this makes them robust against illumination changes.

### 3.4 Gait-based and motion-based models in re-identification

A person's gait is a biometric that seems useful for re-identification because it is difficult for people to deliberately alter the way they walk without looking unnatural. The first method to demonstrate that humans could be recognised by their gait was introduced in the 1970s. The method employed a moving light display [82]. Lights were fastened to the major joints of the human body, and only the lights were visible as the person moved in complete darkness. Analysing the frames showed that the person could be recognised via this pattern. A person's gait changes with walking speed, type of clothing and even mood. However, these factors can be considered as constant within a short period of the re-identification process. The approaches that have used gait analysis are mostly used for recognition and identification purposes, and rarely used for re-identification. In identification, the processing time is not important because the procedure is offline, whereas in re-identification the processing time is an important factor. Therefore gait recognition techniques that require extensive computations are not suitable for use in re-identification.

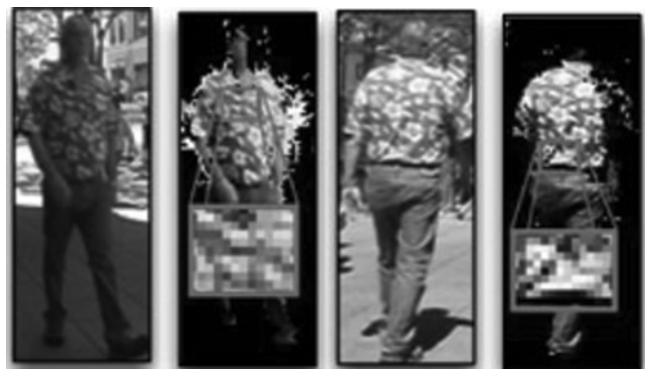

**Fig. 12** *RHSP textural structures [16]*

The drawback of gait-based methods is that the subject must be observed for at least one or two steps before an analysis can be done. Mostly, the gait is extracted from the side view of a person. Thus, the gait-based algorithms require a pure side view of individuals to extract the gait features, which limit these methods. However, to extract gait features, images with high resolution are not needed, which is a benefit of gait-based approaches.

Two types of methods are used in gait recognition: model-based and motion-based. In the model-based approach, a model is fitted to the video data and the parameters of the model are used for identification. This approach is computationally expensive and time consuming because of the large number of parameters. It similarly involves problems such as determining the positions of the joints in the arms and legs. In contrast, motion-based approaches directly use the binary information of sequences using their gait and motion information. The quality of the frames and their resolution is not important in these approaches. Thus, a motion-based approach is considered more often in gait-based methods. Motion-based methods use actual images of a subject during a walking sequence, and extracts features from them. Some features considered are the silhouette and contour of the person. The viewpoint affects the functioning of motion-based methods. Moreover, the high dimensionality of the feature vectors may cause problems (referred to as 'the curse of dimensionality'). On the positive side, motion-based methods are cheaper and easier to calculate, and are less technical to implement [10].

The method in [10] attempts to transform the extracted features of the silhouettes in frames to provide a gait representation based on the sequences of the silhouettes of the subject, which are then used in re-identification by comparing the representations from different silhouettes using a simple classification method. The methods used in this approach are: (i) the active energy image (AEI) [83] and gait energy image (GEI) representations, (ii) the 3D Fourier transform [84] of the gait silhouette volume (to remove high-frequency noise from silhouettes), (iii) the frame difference energy image, (iv) the self similarity plot [85] and (v) a method that uses the distance curves of a sequence of contours. If $\{B_t(x, y)\}_{t=1}^{N}$ are sequences of silhouettes the GEI is defined as

$$G(x, y) = \frac{1}{N}\sum_{t=1}^{N} B_t(x, y) \qquad (8)$$

In this grey-level image, the frequently appearing regions of silhouettes will become brighter. The definition for AEI ($A_t$) is as follows

$$D_t(x, y) = \begin{cases} B_t, & \text{if } t = 1 \\ B_{t-1}(x, y) - B_t(x, y), & \text{if } t > 1 \end{cases} \qquad (9)$$

$$A_t(x, y) = \frac{1}{N}\sum_{t=1}^{N} D_t(x, y) \qquad (10)$$

Fig. 13 shows GEI and AEI images.

In these approaches, the re-identification procedure is regarded as a pipelined process that starts with distinguishing the interesting parts of the video (i.e. the people) from the uninteresting stationary background. This result is achieved using a mixture of Gaussians algorithm. Using the segmented video, the positions of different people are tracked as they move. The position data are then used together with the segmented video frames to create a representation of the different persons' gaits. Subsequently, re-identification is performed by comparing the different gaits using a simple classification procedure (nearest neighbour). In this approach, the mixture of Gaussian distributions [34] is used in modelling each background pixel, and an effective estimation of the parameters based on an expectation maximisation approach [86] is done for the segmentation of interesting objects (people).

In contrast to the interest point operator approach in [11], which generates a large number of potential correspondences, model-based algorithms (which represent the second approach used in that particular study) establish a map from one individual to another. Specifically, a decomposable triangulated graph [87] is used to model the articulated shape of a person as is depicted in Fig. 14. This method can be categorised in the model fitting category of the gait-based methods. A dynamic-programming algorithm is used to fit the model to the person's image [87]. Model fitting localises different body parts such as the arms, torso, legs and head, thus facilitating the comparison of the appearance and structure between corresponding body parts. The main shortage of this modelling is that it works on front view of the gaits which in real scenarios many frames are not from the front view.

Kawai *et al.* [32] proposed a spatiotemporal HOGs (STHOGs) as a gait descriptor to extract both shape and motion features of the silhouettes. In this descriptor, the spatial and temporal gradients of two subsequent frames are calculated for an ($x$, $y$, $t$) space, that is: $\boldsymbol{G} = [G_x, G_y, G_t]$. Then, the orientation of the spatial ($\phi$) and temporal ($\theta$) gradients is calculated

$$\varphi = \tan^{-1}\left(\frac{G_y}{G_x}\right), \quad \theta = \tan^{-1}\left(\frac{G_t}{\sqrt{G_x^2 + G_y^2}}\right) \qquad (11)$$

The temporal and spatial orientations are quantised separately into 9 bins each and then combined into a single 18-bins histogram to construct STHOG. Since this descriptor is so sensitive to the contour of the foreground, removing the whole background is not suitable. Therefore a background attenuator [88] is used instead. Finally, this gait feature is combined with colour features (holistic HSV histogram of silhouettes) to form a mixture of gait and colour features. The problem with this descriptor is that it is too sensitive to differences in the point of view.

Pose energy image (PEI) is another gait feature [14] that is used for re-identification. First, the gait cycle is divided into $K$ different poses. Then, the averages of all the silhouettes belonging to one person in a particular pose of $k_i$ are calculated to obtain $K$, PEIs. An unsupervised $K$-means clustering is used to classify each gait frame into one of the $K$ poses. When all the frames in a sequence are allocated to $K$ poses, the fraction time of the occurrence of each $k_i$ in a

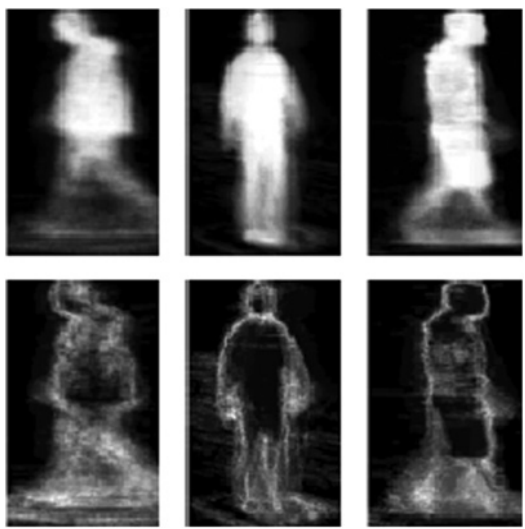

**Fig. 13** *Depiction of GEI (first row) and AEI (second row) images* [10]

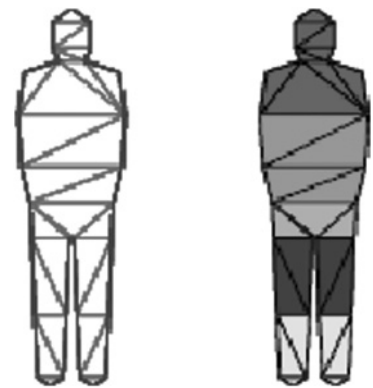

**Fig. 14** *Decomposable triangulated graph used as person model* [11]

gait cycle for $N$ frames is estimated as follows

$$T_i = \frac{1}{N} \sum_{t=1}^{N} \begin{cases} 1, & \text{if frame } f_t \text{ belongs to } k_i \\ 0, & \text{otherwise} \end{cases} \quad (12)$$

If the binary silhouette of frame $t$ in the sequence is $I_t(x, y)$, the $i$th $\text{PEI}_i$ is defined as follows

$$\text{PEI}_i(x, y) = \frac{1}{N \times T_i} \sum_{t=1}^{N} I_t(x, y), \quad (13)$$

if $I_t(x, y)$ belongs to $k_i$

This gait descriptor is more robust than other gait descriptors like GEI and AEI because it contains finer temporal information of the gait and shows exactly how the shape of a person changes. However, as other gait features without temporal information (sequence availability) this descriptor cannot be formed.

### 3.5 Classifiers and matching metrics in re-identification approaches

The third stage of re-identification is to compare the extracted features to find the most similar correspondents. Euclidean distance and Bhattacharyya coefficients are the most popular distance metrics in re-identification [63]. The general form of the Euclidean and Bhattacharyya distance measures is as follows

$$\text{Euclidean dist}(a, b) = \|a - b\|,$$
$$\text{Bhattacharyya coeff}(p, q) = \sum \sqrt{pq} \quad (14)$$

where operand $\|\cdot\|$, is the norm of two vectors. For the Euclidean distance, $a$ and $b$ represent the feature vectors of any dimension and for the Bhattacharyya coefficients $p$ and $q$ represent two different probability distributions of features. As colour and texture histograms are probability distributions, this metric is widely used to compare such features in corresponding samples. In some cases, the formulation of Bhattacharyya $\text{dist}(p, q) = \sqrt{1 - \text{Bhattacharyya coeff}(p, q)}$ is used to better represent properties of a metric structure. It must be noted that when the features do not belong to the Euclidean space, the Euclidean distance cannot be used. For example as previously stated, the covariance descriptors do not lie in the vector space. Thus, special covariance distance metric is used for these descriptors

$$\rho(C_i, C_j) = \sqrt{\sum_{k=1}^{d} \ln^2 \lambda_k(C_i, C_j)} \quad (15)$$

where $C_i$ is a covariance descriptor of dimension $d$ with generalised eigenvalues of $\lambda_k(C_i, C_j)$. The sum of the quadratic distances [74], the sum of absolute differences [3, 58], the correlation coefficients [51] and the Mahalanobis distance [16, 52] are other metrics that are also being used to compare the feature vectors to finally find the most similar individuals in different scenes. The advantage of Mahalanobis distance compared with the other previously mentioned distant metrics is that it counts the correlation between the feature vectors. Although the aforementioned metrics are different and have their own special definitions, they have one major concept in common, which is their restrictive nature and non-flexibility. In other words, they treat any feature that is fed to them equally and do not have the ability to discard useless features. In re-identification, this property may present a large limitation for these metrics. Under severe changes in the illumination, pose and view point conditions, some features may be more distinctive than others. Thus, some features must be given more weight and some must be discarded. However, the standard metrics cannot discriminate between the features. Based on this shortcoming, some researchers have recently applied learning distance metrics [62, 89], optimised distance metrics [90] or probabilistic distances [45, 61] to re-identification to overcome this problem. In these approaches, an attempt is made to solve the re-identification from a distance learning point of view.

Zhao *et al.* [91] proposed a method in which only the salient patches of the foreground were selected and compared together. The salient patches were selected based on learning methods. This kind of selection enables the system to only compare the most relative parts together. They examined $k$-nearest neighbour and one-class support vector machine (SVM) to select the salient patches. Based on their experiments, there was no major difference between results of k-nearest neighbour (KNN) and one-class SVM. The one-class SVM was only trained by positive samples and its goal was to detect outliers. The one-class SVM, on the other hand, was formulated as an optimisation problem which defined a hyper sphere in the feature space. The goal was to minimise the objective function while including most of the training samples inside the hyper sphere. The objective function for such problem was as follows

$$\min_{R \in \mathbb{R}, \xi \in \mathbb{R}^l, c \in F} R^2 + \frac{1}{vl} \sum_i \xi_i \quad (16)$$

$$\text{s.t.} \quad \|\Phi(X_i) - c\|^2 \leq R^2 + \xi_i, \quad \forall i \in \{1, \ldots, l\}: \xi_i > 0$$

where $\xi$ represents the misclassification error, $R$ and $c$ represent radius and centre of hyper sphere and $\Phi(X_i)$ is multi-dimensional feature vector of training data $X_i$ with $l$ training samples. In this equation, $v$ is a trade-off parameter which takes a value between 0 to 1. The SVM was used in some re-identification research as classifier [35, 90, 92] in different styles. The common point in all approaches is that an objective function must be optimised and the hyper-plane must be selected in such way that the vectors from two classes can be separated with maximum margins. In [38], to reduce the complexity cost and speed up the training phase, an active learning method [93] was exploited. In this method, only the samples which were near to the decision plane were labelled and used for learning. This helped to reduce the number of samples which have negative effect on the classifier. The training phase started with one positive and one negative sample. After the first round, only the closest samples to the hyper-plane were selected. The specific selection of samples in this manner decreased the training set to 1/4 of the total number of samples.

Gray and Tao [12] and Bak *et al.* [41] used the AdaBoost scheme to construct an ensemble of likelihood ratios to form a similarity function. In [78], a similar boosting algorithm was used to select more discriminative features among all features.

- Given training samples $\{(x_1, y_1), \ldots, (x_L, y_L)\}$, $x_l \in R^m$, $y_l \in \{-1, 1\}$
- Initial uniform weight distribution of $D_1(l) = \frac{1}{L}$ is considered over all samples
- For $t = 1:T$
  - Apply weak classifiers ($h_t$) on data samples with respect to current distribution $D_t$ and choose the one which minimises misclassification error: $err = \sum_{y_l \neq h_t(x_l)} D_t(l)$
  - Set current weight $\alpha_t = \frac{1}{2} \ln \frac{1-err}{err}$
  - Update $D_t$: $D_{t+1} = D_t \exp(-\alpha_t h_t(x_l) y_l)$
  - Add $\alpha_t$ and $h_t$ to the ensemble
- Output final classifier which is combined of T weak classifiers: $H(x_l) = \sum_{t=1}^{T} \alpha_t h_t(x_l)$

**Fig. 15** *AdaBoost algorithm*

In this approach, a subset of *T* more informative features which corresponded to the weak classifiers in the boosting algorithm was selected among the whole set $\{f_1, \ldots, f_M\}$. Fig. 15 shows the procedure of the adaptive boosting method. The main disadvantage of the boosting method is that they can easily be over-fitted and are sensitive to noise and outliers.

Ensemble of decision trees known as random forests [94] is also another tool which has been used for classification in re-identification. The ensemble of decision trees is less sensitive and has reduced variance compared with an individual decision tree. Du *et al.* [95] proposed a model, known as random ensemble of colour features (RECF) in which they used random forests to learn a similarity function $f(.)$ based on different colour spaces channels as features. Given $(x_i, y_i)$ as training samples where $x_i$ denotes a pair of person images and $y_i$ denotes the label for the given $x_i$. The similarity function is a combination of *T* posteriors probabilities $p_t(x_i)$ of *T* trees and is formulated as follows

$$f(x_i) = \frac{1}{T} \sum_{t=1}^{T} p_t(x_i) \quad (17)$$

where $p_t(x_i)$ is the posteriors probability of a leaf node *l* of tree *t* and is estimated based on the fraction of positive subsamples which reach to the leaf node ($N_{l-pos}$) to the total number of subsamples reach to that leaf node ($N_{l-pos} + N_{l-neg}$).

Zheng *et al.* [61, 89] introduced the novel probabilistic relative distance comparison (PRDC) and relative distance comparison (RDC) models, in which they tried to formulate the re-identification as a distance learning problem. In contrast to conventional approaches that have attempted to minimise the intra-class variation (i.e. images of one person) and maximise the inter-class variation (i.e. images of two different persons), the objective function used by PRDC aims to maximise the probability of a pair that is a true match (i.e. two true images of person A) having a smaller distance than that of a pair that is a related wrong match (i.e. two images of persons A and B, respectively). If *f* is considered for PRDC, it must be learned so that the distance between relevant pairs is less than that between irrelevant ones

$$f(x_i^p) < f(x_i^n) \quad (18)$$

where $x_i^p$ is the distance between two relevant pairs, and $x_i^n$ is the distance between two irrelevant pairs. The probability of the above event is computed, and then the function is learned based on the maximum likelihood principle

$$f = \arg \min r(f, \mathbb{O}) \quad (19)$$

$$P(f(x_i^p) < f(x_i^n)) = (1 + \exp\{f(x_i^p) - f(x_i^n)\})^{-1} \quad (20)$$

$$r(f, \mathbb{O}) = -\log \left( \prod_{\mathbb{O}_i} P(f(x_i^p) < f(x_i^n)) \right) \quad (21)$$

$$\mathbb{O} = \{\mathbb{O}_i = (x_i^p, x_i^n)\} \quad (22)$$

This function can also be learned by SVM as was done by Prosser *et al.* [90]. The difference between RDC and RankSVM is that RDC uses a logistic function that makes a soft margin measure for vectors $x_i^n$, whereas RankSVM does not have such a margin. Thus, RDC would be more robust against inter-class and intra-class variations. Large margin nearest neighbour (LMNN) [61, 96] classifier is another kind of learning distance metric that is used for re-identification. In this approach, a linear transformation *L* is learned in order to minimise the distance between a data point and its *k*-nearest neighbours with the same label, while simultaneously maximising the distance of this data point from differently labelled data points. Khedher *et al.* [45] developed two GMM to model the distance distributions of relevant pairs (GMM$_1$) and irrelevant pairs (GMM$_2$). To decide whether the test distance (*d*) is relevant or not, its likelihood ratio must be >1

$$\text{LR} = \frac{P(d|\text{GMM}_1)}{P(d|\text{GMM}_2)} \quad (23)$$

$$P(d|\text{GMM}_i) = \sum_{g=1}^{G} C_{gi} \frac{1}{\sqrt{2\pi\sigma_{gi}^2}} \exp\left(\frac{-1}{2}\left(\frac{d - \mu_{gi}}{\sigma_{gi}}\right)^2\right) \quad (24)$$

where $\mu_{gi}$, $\sigma_{gi}$ and $C_{gi}$ are the mean, variance and weight of component *g* from GMM$_i$.

### 3.6 Datasets for re-identification

There are some standard databases that have been used in different works for the evaluation and comparison of re-identification approaches. Farenzena *et al.* [16] used public databases such as viewpoint invariant pedestrian recognition (VIPeR) [12], imagery library for intelligent detection system (i-LIDS) [97] (which was also used by Zheng *et al.* [61]), and ETH Zurich (ETHZ) [98]. In [27],

Satta *et al.* similarly used VIPeR to compare the results with [16]. Bak *et al.* [41] used the TRECVID database (organised by National Institute of Standards and Technology (NIST)) to train the human detection algorithm with 10 000 positive (human) samples and 20 000 negative (background scene) samples. They similarly used CAVIAR [99] as did Hamdoun *et al.* in [3, 58]. In another work, Bak *et al.* [43] used i-LIDS as their database. Khan *et al.* [59] also used the NICTA and CAVIAR datasets to evaluate their results.

In some studies, datasets were created based on the scenarios in which the re-identification should be done. For instance, Cong *et al.* [2] created real datasets using two cameras in the desired places. Gheissari *et al.* [11] also used their own datasets. Skog [10] and Annesley *et al.* [33] both made two different datasets based on the scenes that they needed.

The frequent use of the above-mentioned datasets in various works has changed them into standard ones for evaluation of experiments in re-identification. They also present challenges that must be overcome for a re-identification. However, they are not appropriate for gait- and motion-based approaches because these approaches require databases that contain successive image sequences with information of their points of views.

The CAVIAR, i-LIDS and TRECVID databases are in a video form, whereas VIPeR contains still images and ETHZ is a set of consecutive frames. Almost all of them contain videos and images of pedestrians from different points of view and with different illumination conditions, but among them only i-LIDS and ETHZ contain occluded frames. TRECVID and VIPeR also have images that show pedestrians carrying objects (partial occlusion). The TRECVID and i-LIDS datasets contain images that were obtained under actual surveillance conditions. Table 2 shows the most discriminatory and challenging features of each of these datasets compared with the others.

### 3.7 Evaluation metrics

One major issue mentioned in the reviewed papers is the way in which their results are evaluated and compared. Most methods have used the cumulative matching characteristic (CMC) curve as stated in [77] to evaluate and compare their results [2, 16, 27, 41, 43, 61]. In CMC curves, the cumulative number of re-identified queries is shown based on the order in which they have re-identified. If the number of true re-identified queries in rank $i$ is $tq(i)$. The amount of CMC for rank $i$ is defined as

$$\text{CMC}(i) = \sum_{r=1}^{i} tq(r) \qquad (25)$$

Other curves also have been used such as receiver operating characteristics [59, 92], precision–recall (PR) [3, 58], and other metrics [10, 11] for evaluation of re-identification systems. In PR curves, the indices are calculated as below

$$\text{precision} = \frac{\text{TP}}{\text{TP} + \text{FP}}, \quad \text{recall} = \frac{\text{TP}}{\text{target number}} \qquad (26)$$

where TP and FP stand for true positives and false positives. There are other numbers of works which prefer PR to CMC curves [6, 55] the only privilege of PR curve to CMC curve is that in PR curve the ratio of false positive re-identified samples can be seen apparently, whereas in CMC it is hidden. However, the lack of the rank in PR curves is a shortage to evaluate the performance of a re-identification system.

Xiang *et al.* [38] evaluated the performance of their re-identification system by measuring the true positive ratio and true negative ratio against different labelled samples in two different curves, but these true ratios are only the first ranked results and these evaluation curves are unable to show the next ranks of re-identified samples.

The privilege of CMC compare with other curves is in CMC not only the first true re-identified query rank is indicated, but also the true re-identified queries in other ranks are also indicated. By this means, the performance of the systems for re-identification can be depicted better. It is important to mention that two important factors in CMC curves are the first rank re-identification rate and the steep of the curve. The steeper the curve the better the performance is.

## 4 Challenges and future direction of people re-identification research

In the literature, the re-identification methods can be grouped into two sets. The first group is composed of single-shot methods that analyse the single image of a person [12, 13]. They are applied in the lack of tracking frames. The second group includes multiple-shot approaches; they employ multiple frames of a person (usually obtained via tracking) to make the signatures [3, 11, 100]. Both approaches simplify the problem by adding temporal reasoning to the spatial layout of the monitored environment to prune the candidate set to be matched [100].

In most security-related research, the colour and/or texture of the clothes are/is regarded as the most significant cues for person retrieval [2, 11, 16, 52] which shows that clothing information can be used for local descriptors. Using other features such as facial and gait features [10, 11, 25] or even the height of individuals [6, 51] have a great advantage because these features are likely to remain constant over a longer period of time. The person's face and gait are popular descriptors in recognition tasks, which have been used even for re-identification [7, 10], but facial-based descriptors tend to produce less accurate results when employed in combination with low-resolution cameras. Nevertheless, those cameras are widely used in actual surveillance systems mainly because of economical reasons.

**Table 2** Most challenging and discriminatory features of each of datasets

| Dataset | CAVIAR | ETHZ | VIPeR | i-LIDS | TRECVID |
|---|---|---|---|---|---|
| challenging features | real situation, low resolution | illumination variation, occlusion | pose variation, background variation | real situation, occlusion | real situation, Partial occlusion |

**Table 3** Re-identification rate for different normalisation methods [101]

| Normalisation method | RGB colour space | Grey world | Histogram equalisation | Affine normalisation |
|---|---|---|---|---|
| first rank of CMC re-identification rate | 70 | 95 | 97.5 | 95 |

As previously mentioned, the three most important existing issues in re-identification are illumination changes, view point and pose variations and scene occlusion. Although insufficient attention has been given to the scene occlusion issue in many of re-identification studies, there are only a small numbers that have provided solutions for it, almost all of them have given attention to illumination and pose variations. The various colour, texture, shape and gait models and descriptors that have been proposed in these works have a common goal, which is to improve the true re-identification rate by overcoming these issues as well as they can. Here, the various solutions given in the literature for these issues will be summarised.

### 4.1 Robustness against illumination variations

Since some colour spaces like HSV and LAB are more robust against illumination changes, some approaches have preferred to use them in histogram descriptors or textural features instead of the RGB colour space [11, 16, 27, 63]. Reducing the number of quantisation levels of colour channels also decreases the sensitivity to noise and intensity [63]. In methods that use colour histograms, a common way to make them illumination invariant is histogram equalisation [2, 51, 74]. The assumption in the histogram equalisation process is that although illumination changes make the sensors respond differently, their rank ordering will remain unchanged. The rank measure for level $i$ is defined as

$$M(i) = \frac{\sum_{t=1}^{i} H(t)}{\sum_{t=1}^{N_b} H(t)} \quad (27)$$

where $N_b$ is the total number of quantisation levels and $H(.)$ is the histogram. The Greyworld and Affine normalisations have also been used. These normalisation methods are being applied on colour channels ($I_k$) instead of histograms

$$\text{norm}_{\text{Grey}}(I_k) = \frac{I_k}{\text{mean}(I_k)},$$
$$\text{norm}_{\text{Affine}}(I_k) = \frac{I_k - \text{mean}(I_k)}{\text{std}(I_k)} \quad (28)$$

where $std$ is the standard deviation of the pixels in colour channel $I_k$. Histogram normalisation is another method used in some works [37, 58]. Cong et al. [101] performed the same algorithm on the data that were grabbed from two locations, but applied different normalisation methods on it. The re-identification rates are as shown in Table 3.

The light absorbed by the camera's sensor is reflected from different sources in the scene. Thus, to obtain more realistic illumination compensation, the influence levels of these local light sources on the object's colours must be indicated. Monari [49] divided scene lights into the three categories of ambient light, overhead light and backlight.

They proposed a model for an object's pixel intensity $I_{\text{object}}(x)$ and its shadow pixel intensity $I_{\text{shadow}}(x)$ which was made up of ambient and overhead lights.

$$I_{\text{shadow}}(x) \equiv K_c \big(I_{\text{amb}}(x) + I_{\text{overhead}}(x)\big) \quad (29)$$

By dividing the shadow intensity by a known ground floor reflection coefficient ($K_c$), the pure light sources were obtained $(I_{\text{amb}}(x) + I_{\text{overhead}}(x))$ and then the object's pixel intensity $I(x)$ was normalised by this value

$$I_{\text{norm}}(x) = \frac{I(x) K_c}{I_{\text{shadow}}(x)} \quad (30)$$

To guarantee the correct illumination compensation in this approach, it is necessary to properly detect the shadows of the individuals and the environmental information of the scene (ground floor reflection coefficient), which are the drawbacks of this method. Aziz et al. [55] classified people appearances into frontal and back appearances to obtain robustness against illumination changes. They also normalised SIFT descriptor that they used for feature extraction.

### 4.2 Robustness against view point and pose variations

One of the most effective methods to obtain pose invariance was devised by Farenzena et al. [16]. In their method, a silhouette is divided into the three main parts of the head, torso and legs. Then, for each part, a symmetry axis divides the silhouette into two parts. Features are then selected from the two sides of this axis and weighted based on their distance from this axis. This symmetric selection of patches to extract features makes it pose invariant. In contrast, the arbitrary and random selection of patches from the torso or legs to extract features from them would make the method vulnerable to pose variations [27]. The rotation-invariant nature of Gabor and Schmid descriptors is the reason for their same response when applied to different poses of the silhouettes and results in the pose invariant features [12]. The distance metrics can also affect the pose invariance of the methods like Mahalanobis distance that was used by Martinel and Micheloni [60] to measure two extracted SIFT descriptors.

Most of the pose variations in different frames occurred around the vertical axes of the scenes. To define the models and descriptors to be $x$ axes independent partially improves robustness against pose variations [12, 51, 61].

### 4.3 Robustness against scene occlusion

The descriptors proposed by Zheng et al. [20] are most relative ones which have been designed to deal with scene occlusions. The descriptors are designed to perform on groups of individuals instead of the single ones, but according to its formulation which is based on inter people distances the method is unable to re-identify the individuals separately.

Wang et al. [102] trained LMNN-R classifier with real and synthesised occluded dataset to overcome the partial occlusion of the test dataset. An interesting body print was proposed in Albiol et al. [51] that was totally robust against occlusions. This body print is extracted from all RGB values of the pixels at height '$h$' (from the ground) of the foreground. The signature uses temporal information of consequent frames of one person. During these consequent frames, the signatures of the occluded frames are being neglected. Thus, the occluded frames do not affect the final signature that is extracted from the whole frames. The height information in this method was grabbed by Microsoft kinect sensor and cannot be used in outdoor environment that is considered as a large limitation for re-identification application. The gait-based approaches are so sensitive against occlusions because even small partial occlusion in consequent frames does not let to extract the gait feature properly. Roy *et al.* [14] used the advantage of the phase of the motion of the silhouettes in a hierarchical way. This feature was used in the frames in which the gait was affected by occlusion.

### 4.4 Real-time re-identification

The ultimate goal for a generic re-identification system is to be capable of working in real-time situation. To do so, the whole re-identification procedure must be fast enough to allow for real-time implementation. Although some attempts for real-time re-identification were noted [35, 37, 56]; there still exist many issues that must be solved before a complete real-time re-identification system can be successfully implemented.

The FSCH descriptor that was exploited by Xiang *et al.* [37] took 1.14 ms to be extracted from a $70 \times 25$ image patch. This algorithm must perform 42 additional computations compared with the simple histogram method. This was mainly because of membership degree and 5D space computations involved. They also performed similarity matching between the extracted descriptors by correlation metric which is quite fast. Goldmann [35] used simple features like RGB value, colour structure descriptor, co-occurrence matrix and intensity-based histogram to be able to implement the real-time system. However, since their method was based on supervised learning method in classification stage it needed a training phase which must be executed prior to the testing phase.

Eisenbach and Kolarow [52] adopted a person detection method based on contour cues [53] and a real-time tracking method [54] to track people to speed up their algorithm. Besides that, they extracted appearance features from upper and lower bodies of any person who passed the camera while recording the video to reduce the total computation time. An online feature selection scheme was used in which the joint mutual information [103] estimated the dependencies between each of the features and the class label of the detected person. Thus, the best features could be selected for specific class and redundant features will be removed. Consequently, the re-identification could be performed faster. In [77], Wang *et al.* used the integral computations to build their occurrence matrix as their descriptor to speed up their algorithm performance. The computation complexity is independent of the size of the rectangular domain D on which the integral computation is applied.

Satta *et al.* [40] reduced the matching time between probe and the gallery images by transposing them into dissimilarity space. The components from the same parts of individuals in the gallery set were put together and clustered to some prototypes. Thus, each part (torso or leg) in the gallery set had its own prototype bank. Then, the difference between each part of every individual in the gallery and centroid of the prototypes was computed and set as dissimilarity vector for each individual. The same action was done for query. In the last step, the nearest individual to the query would be selected based on lowest distance between dissimilarity vectors. The matching time in their experience was <0.01 ms. In this method, instead of any measure to compare complex descriptors only the vectors were compared based on simple distance metric Hausdorff distance [104] which drastically reduced the computation cost and memory usage.

To implement a re-identification system in real-time, the background subtraction, human detection, feature extraction and matching steps must all be based on real-time algorithms. However, the methods mentioned in here have only partially implemented their algorithms in real-time. As such, a completely real-time re-identidifcation system is very much needed, and thus real-time re-identification research will be an open issue among the researchers.

Table 4 lists the most popular features and descriptors used in state-of-the-art re-identification works which have been mentioned in the previous sections. The features and descriptors are namely colour, texture, shape and gait. In this table, these features and descriptors are compared based on their effectiveness in solving re-identification issues and a qualitative comparison between these features is also provided. The robustness of these features and descriptors against varying illumination, pose and occlusion are shown. In addition, the acronyms 'CI' and 'PI' in Table 4 are short for 'completely independent' and 'partially independent', respectively. For cases in which the invariance is dependent of some other factors, the acronym 'DI' which stands for 'dependent invariance' is used.

Actually, the major significant function of pure colour features and descriptors is to make signatures that are invariant against varying illumination. As stated in Section 4.1, several methods have been used to make these signatures independent against illumination changes, but to make the signatures pose invariant they must be combined with complementary spatial and textural features. The methods which have used region-based colour representations definitely outperform the holistic colour representations. Table 5 represents an example of the first rank of CMC re-identification rates of spatial covariance descriptor (SCR) proposed by Bak *et al.* [43] and a normal colour histogram with and without using group context model (CRRRO and BRO) proposed by Zheng *et al.* [20]. The i-LIDS dataset was used for these experiments.

As can be seen from the table, the colour representation without spatial information has the lowest re-identification rate. However, when this colour representation is combined with the spatial group context models which are CRRRO and BRO descriptors the results are significantly improved. As indicated, the SCR descriptor outperforms both because the covariance matrices which are extracted from the overlapped regions are robust against pose and partial occlusions. The first derivative of grey level channel that is utilised in constructing covariance feature vectors has also made it illumination invariant and improved the re-identification rate.

In addition, interest point descriptors are also attractive tools to use in re-identification. Nevertheless, the reported results about their performance are somewhat varied. In [45, 92], it was reported that SURF outperformed the SIFT, but in [72] SIFT and GLOH outperform SURF and CCPD. However, the datasets in which these descriptors were applied and the methods used in segmentation and classification processes were known to have significant effect in the final re-identification rate. In applying interest point descriptors, the important queue is to decrease the number of corresponding interest regions to control the computational costs. The advanced methods that their bases are related to interest point descriptors like visual words or attribute-based methods are new approaches in re-identification which can also direct the future aspects of re-identification [105, 106].

Table 4  Summary of the various main features and descriptors used in re-identification

| Research which used feature/descriptors | Feature/descriptor names | Types of feature/descriptor | Illumination invariances | Pose invariances | Occlusion invariances | Remarks |
|---|---|---|---|---|---|---|
| Gheissari et al. [11], Satta et al. [27], Wang and Lewandowski [102] | colour histograms | colour | DI | × | × | using HSV and LAB colour spaces make histograms partially robust against illumination changes, but they are totally vulnerable against pose differences and occlusion |
| D'Angelo and Dugelay [63] | PCH | colour | PI | × | × | — |
| Xiang et al. [37] | FSCH | colour | PI | DI | × | the x dimension of the descriptor must be ignored for pose invariance |
| Annesley et al. [33], Bak et al. [41] | MPEG7 | colour | DI | PI | × | DCD descriptor is not pose invariant, but colour layout descriptor (CLD) descriptor is pose invariant |
| Khan et al. [59] | CCPD | colour | PI | PI | PI | the polar representation of the descriptor makes it pose invariant the type colour space is important for illumination invariance |
| Farenzena et al. [16], Satta et al. [27] | RHSP | texture | × | CI | × | — |
| Zhang et al. [19], Gray and Tao [12] | Gabor and Schmid | texture | CI | CI | × | — |
| Zhang et al. [19], Corvee et al. [48], Hirzer et al. [62] | LBP | texture | DI | × | × | LBP is invariant to grey level channel |
| Goldmann [35] | co-occurrence matrices | texture | DI | CI | × | this descriptor must be applied on grey level colour channels in order to be robust against varying illuminations |
| Gheissari et al. [11] | frequency image | gait | PI | PI | × | — |
| Skog [10] | AEI | gait | PI | CI | × | — |
| Skog [10] | GEI | gait | PI | CI | × | — |
| Kawai et al. [32] | STHOG | gait | CI | PI | × | contains shape and motion features |
| Roy et al. [14] | PEI | gait/shape | CI | DI | × | the side view of silhouette must be available for PEI extraction |
| Bauml and Stiefelhagen [72] | SIFT | colour, texture | PI | PI | PI | — |
| Khedher et al. [45], Hamdoun et al. [58] | SURF | colour, texture | PI | PI | PI | — |
| Bauml and Stiefelhagen [72] | GLOH | colour, texture | PI | PI | PI | — |
| Zhang et al. [19], Bak et al. [43, 75] | covariance matrices | colour, texture | PI | CI | PI | the covariance matrices must be extracted from overlapped regions in order to be robust against occlusion |
| Zheng et al. [20], Xiang et al. [38] | HOG | colour, texture | DI | CI | PI | the training set plays salient role to make the descriptor robust against pose and occlusion |
| Zheng et al. [20] | CRRRO and BRO | colour, texture | DI | CI | CI | the selection of colour feature to construct the descriptor is important for its illumination invariance |

On using gait features in re-identification, the advantage is that the low resolution of CCTVs does not affect them, but the main disadvantages of gait features are (i) their susceptibility to occlusion, that is, the gait cannot be captured properly if the silhouette is occluded by other objects or the silhouettes of other people and (ii) their need for a side camera-view because most of the gait features can be defined based on the side camera-view of the gait. These shortcomings alongside with high-computational demand for gait features make them less favourable in re-identification approaches. However, the gait-based features can be very effective supplements in companion with appearance-based models. The hierarchical models which are designed to fuse the appearance and gait/motion features have proven to be useful and can be applied in forthcoming research on re-identification [14, 21]. To have a quantitative comparison of the above-mentioned statement, the results presented by Kawai *et al.* [32] are shown again in Fig. 16 in which comparisons between the first rank re-identification rate of using their proposed STHOG gait descriptor method, a HSV colour histogram as pure colour feature method and the fusion of the two features are made. The descriptors were applied on their own dataset with different viewing angles. The query viewing angle is near to ID5 which indicates the side view. As shown in Fig. 16, in the viewing angles near to the query image the gait feature outperforms the colour feature (ID4, ID5 and ID6), but when the pose starts to differ, the gait feature is unable to outperform the colour feature (ID1, ID2 and ID3). However, in all viewing angles the fusion of gait and colour features produced better results than using them alone which is somewhat expected. The point to highlight is that in this graph, one can see the performance of the descriptors for different viewing angles which has never been reported previously.

**Table 5** Comparative results of using holistic pure colour features against using colour features in combination of spatial features [43]

| Method | Colour histogram without group context (CRRRO and BRO) | Colour histogram with group context (CRRRO and BRO) | SCR |
|---|---|---|---|
| first rank re-identification rate of CMC curve, % | 10 | 17 | 33 |

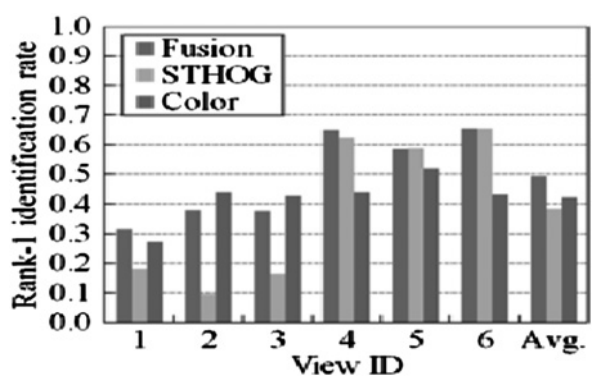

**Fig. 16** *Re-identification rate for STHOG gait descriptor, HSV colour histogram and combination of them [32]*

**Table 6** First rank re-identification rate of learning metrics and Bhattacharyya coefficients for different datasets [61, 89]

| Learning metrics/classifier | First rank re-identification rates | | |
|---|---|---|---|
| | VIPeR | i-LIDS | ETHZ |
| RDC | 15.66 | 44.05 | 72.65 |
| AdaBoost | 8.16 | 33.58 | 69.21 |
| LMNN | 6.23 | 33.68 | 64.88 |
| Bhattacharyya distance | 4.65 | 31.77 | 60.97 |

The application of learning distance metrics in re-identification is intensely growing and it is anticipated that it will become a prevalent area of research in the near future. In fact, the strategy of these metrics which is to maximise the inter-cluster distance while minimising the intra-cluster distance has performed well especially on appearance-based models which suffer from the noise and variations even at the same clusters.

Table 6 is a comparative evaluation of the first ranks of CMC curves of different learning metrics and Bhattacharyya distance discussed on previous sections for VIPeR, i-LIDS and ETHZ datasets. These presented results were obtained from Zheng *et al.* [61, 89]. The same colour and texture features were extracted from the persons of interest and then the feature vectors were fed to these different classifiers. For training, the number of pedestrians as samples were 316, 40 and 30 in VIPer, ETHZ and i-LIDS, respectively.

As expected, the all other learning metrics outperformed the Bhattacharyya distance metric which is shown in Table 6. The RDC was the best learning metric and has produced the best results of 15.66, 44.05 and 72.65% for the VIPeR, i-LIDS and ETHZ datasets, respectively. It is worth mentioning that the RECF [95] has 16.96% accuracy rate on VIPeR which is above RDC and shows its efficiency to alleviate over-fitting problem when there are limited number of training data. However, this classifier was not examined on the other existing datasets. The challenging VIPeR dataset, as expected has the lowest re-identification rates when compared with the results using i-LIDS and ETHZ datasets. Obviously, the learning metrics can perform better on video databases of ETHZ and i-LIDS when compared with VIPeR. In terms of performance, the reason behind RDC high accuracy is that the RDC or also known as PRDC learning metric adopts the second-order moments of features and consequently the joint effects of the features are considered, whereas the other classifiers assume each feature independently. Therefore there is no interaction between the features. Performance in terms of the datasets used, ETHZ and i-LIDS were better because of the consequential nature of the frames in them, whereas the VIPeR dataset only consists of single shots.

**Table 7** Highlighted methods which are capable of being pursued in future research on re-identification

| Re-identification stages | Highlighted methods | Examples | Key points |
|---|---|---|---|
| feature extraction stage | 1. attribute extraction | Layne *et al.* [105] | ability to more meaningful representation of objects |
| | 2. interest point detectors | Bauml and Stiefelhagen [72], Martinel and Micheloni [60] | high level of robustness against illumination and pose variations |
| | 3. fuzzy analysis | Xiang *et al.* [37], D' Angelo and Dugelay [63] | ability to handle severe illumination changes |
| | 4. spatiotemporal methods | Kawai *et al.* [32], Bedagkar-Gala and Shah [47] | simultaneously uses temporal, spatial, colour and sometimes gait of the frames |
| classification stage | 1. distance learning metrics | Zheng *et al.* [89], Mignon and Jurie [96] | makes classifier to be able to be discriminative and to be optimised against different features |
| | 2. feature selection methods | Hirzer *et al.* [78] | reduce computation cost in classification, most relevant features will be selected |

Generally, in re-identification the models must not be only descriptive, but they have to be discriminative simultaneously. Features extracted from the silhouettes may construct completely descriptive descriptors, but most of these features are not discriminative enough. The learning distance methods try to move the load of discrimination from the feature to the classifier. This property is rarely found in methods with usual distance classifiers [3, 58, 63].

Finally, we also recommend and highlight several potential approaches to extract features and perform classification for re-identification which are summarised in Table 7.

## 5 Summary

In this paper, we provide a review of the existing state-of-the-art research on re-identification which involves both the appearance and gait/motion descriptors and spell out the abilities, limitations and advantages of the various available methods for re-identification. We begin the paper by discussing the issues pertaining to people re-identification which involves inter- and intra-camera issues. Among them, the most serious issues include illumination, pose changes and scene occlusions. Next, we afford methods that have been used for person re-identification in which different types of descriptors that are based on colour, texture and gait of the silhouettes are described. The different classifiers, the existing standard datasets and evaluation methods for re-identification are also explained.

We also highlight the challenges that need to be resolved which mainly concerns the robustness against the following aspects of illumination, view point and pose variations and scene occlusion. Although several solutions of the above-mentioned issues have been proposed and developed, the existing methods are still unable to overcome those issues completely. Truthfully, the solutions thus far, are not applicable in all practical scenarios in which illumination variations, pose changes and occlusions may occur simultaneously with respect to time and if these were to occur simultaneously, it will incur a high-computational cost. As such, new and improved descriptors and models are needed so that the issues can be solved more efficiently.

In this paper, we have provided both qualitative and quantitative comparisons of several re-identification methods to depict the advantages and shortcomings of the models being used. Finally, to provide an insight for the future research direction, we highlight the methods that are capable of being pursued in the forthcoming research.

To conclude, we believe that research in re-identification will proliferate as the demand for efficient intelligent video surveillance system increases so as to ensure secured and safe environment of the society and mankind.

## 6 Acknowledgments

The authors would like to express their gratitude to the Malaysian government and the Universiti Kebangsaan Malaysia for providing financial assistance via grant LRGS/TD/2011/UKM/ICT/04/02 and DPP-2013-003 given for this project.